\documentclass[runningheads]{llncs}
\usepackage[T1]{fontenc}
\usepackage{graphicx,verbatim}
\usepackage{algorithm}
\usepackage{algorithmic}
\usepackage{bbding,pifont}
\usepackage{mathtools} 
\usepackage{amsmath,amssymb}
\usepackage{enumitem}
\usepackage{graphicx}
\usepackage{subfig}
\usepackage{siunitx}
\usepackage[colorlinks,linkcolor=red]{hyperref}
\usepackage{color}
\usepackage[noadjust]{cite}
\usepackage{xcolor}
\usepackage{pifont}
\usepackage{lineno,hyperref}
\usepackage{booktabs}
\usepackage{multirow}
\usepackage{booktabs}

\begin{document}
\title{Topology-Driven Transferability Estimation of Medical Foundation Models for Segmentation}

\titlerunning{Topology-Driven Transferability Estimation}
%
\author{Jiaqi Tang\inst{1,4,5,6}\thanks{These authors contributed equally to this work.} \and
Shaoyang Zhang\inst{2}\protect\footnotemark[1] \and
Xiaoqi Wang\inst{3} \and
Jiaying Zhou\inst{1} \and
Yang Liu\inst{1} \and
Qingchao Chen\inst{1,4,5,6}\Envelope
}

\index{Tang, Jiaqi}
\index{Zhang, Shaoyang}
\index{Wang, Xiaoqi}
\index{Zhou, Jiaying}
\index{Liu, Yang}
\index{Chen, Qingchao}

\authorrunning{J. Tang, S. Zhang et al.}

\institute{Peking University, Beijing, China \and
Hohai University, Nanjing, China \and
Beijing Normal University-Hong Kong Baptist University United International College, Zhuhai, China \and
National Institute of Health Data Science, Peking University, Beijing, China \and
Institute of Medical Technology, Peking University, Beijing, China \and
State Key Laboratory of General Artificial Intelligence, Peking University \\
\email{\{jiaqi\_tang, qingchao.chen\}@pku.edu.cn, zsyhhu04@163.com} \\
\Envelope\ Qingchao Chen is the corresponding author}

\maketitle              
\begingroup
\renewcommand\thefootnote{}
\footnotetext{Accepted at MICCAI 2026.}
\endgroup

\begin{abstract}
The advent of large-scale self-supervised learning (SSL) has produced a vast zoo of medical foundation models. However, selecting optimal medical foundation models for specific segmentation tasks remains a computational bottleneck. Existing Transferability Estimation (TE) metrics, primarily designed for classification, rely on global statistical assumptions and fail to capture the topological complexity essential for dense prediction. We propose a novel Topology-Driven Transferability Estimation framework that evaluates manifold tractability rather than statistical overlap. Our approach introduces three components: (1) Global Representation Topology Divergence (GRTD), utilizing Minimum Spanning Trees to quantify feature-label structural isomorphism; (2) Local Boundary-Aware Topological Consistency (LBTC), which assesses manifold separability specifically at critical anatomical boundaries; and (3) Task-Adaptive Fusion, which dynamically integrates global and local metrics based on the semantic cardinality of the target task. Validated on the large-scale OpenMind benchmark across diverse anatomical targets and SSL foundation models, our approach significantly outperforms state-of-the-art baselines by around \textbf{31\%} relative improvement in the weighted Kendall's $\tau$, providing a robust, training-free proxy for efficient model selection without the cost of fine-tuning.
The code will be made publicly available upon acceptance.

\keywords{transferability estimation  \and transfer learning \and foundation model selection}
\end{abstract}

\section{Introduction}
\label{sec:intro}

\begin{figure*}[t]

    \includegraphics[width=0.36\linewidth, height=0.37\linewidth]{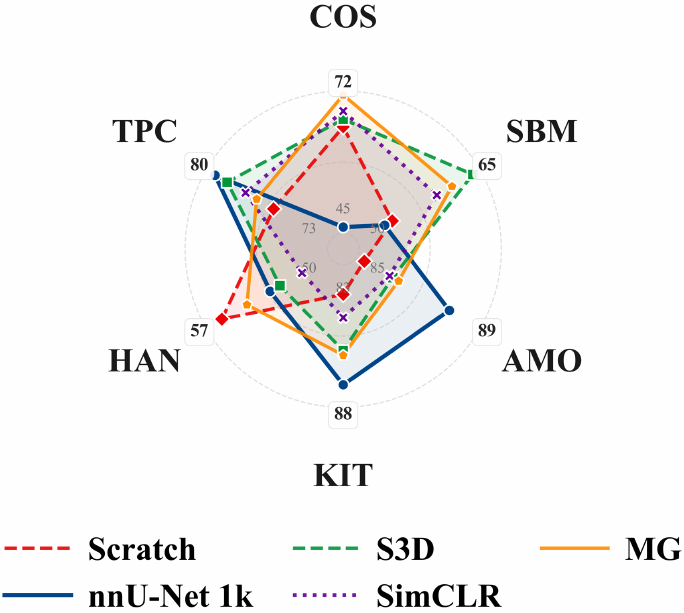} 
    \includegraphics[width=0.62\linewidth, height=0.37\linewidth]{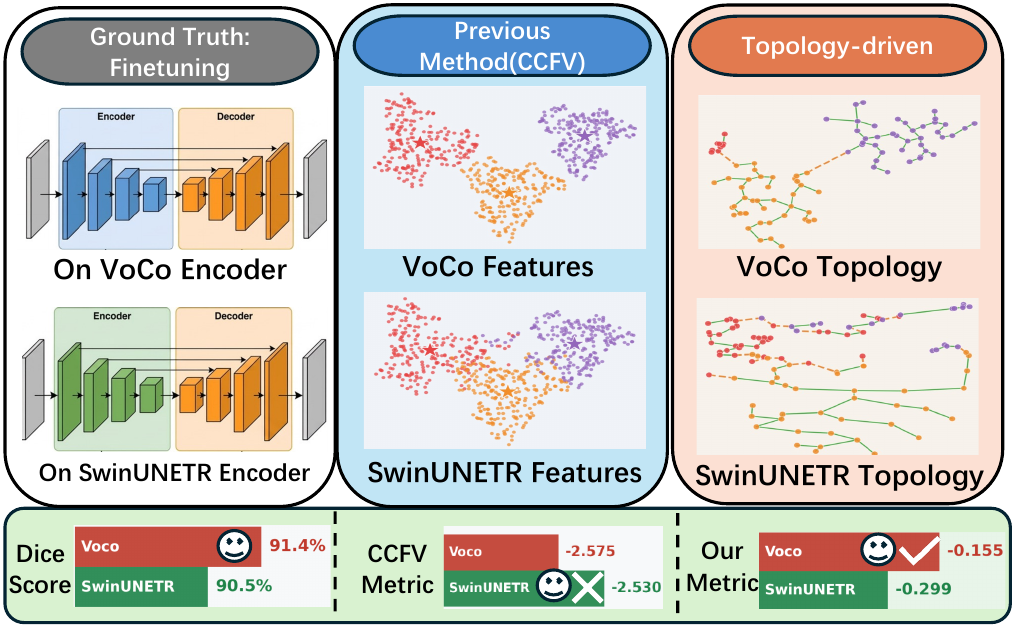} 

    \caption{ \textbf{Left:} Fine-tuning performance of diverse foundation models varies across downstream datasets, indicating that the optimal pre-trained encoder is task-dependent. \textbf{Right:} Visualizing transferability: Statistics vs. Topology. We compare the ground truth fine-tuning performance (Left) against rankings from the statistical metric (Middle) and our topology-driven metric (Right). }
    \label{fig:teaser}
\end{figure*}

The rise of large-scale self-supervised learning (SSL) has created a diverse zoo of medical foundation models trained on massive unlabeled data to learn general-purpose representations~\cite{chen2020simple,zhou2021models,He2022MAE,chen2023masked,Wu2024Voco,Tang2022SwinUNETR}. Yet, the best pre-trained encoder is highly task-dependent: as illustrated in Fig.\ref{fig:teaser} (left), different downstream segmentation datasets favor different source models, making exhaustive fine-tuning a costly combinatorial search\cite{wald2025openmind}. This motivates a training-free Transferability Estimation (TE) framework that predicts post-fine-tuning segmentation performance directly from pre-training features, enabling fast and resource-efficient model selection.

Existing TE metrics, largely developed for image classification~\cite{nguyen2020leep,you2021logme,pandy2022transferability,yang2023pick,tan2021otce,ding2022pactran}, can be misaligned with the “SSL encoder + segmentation decoder” setting. Methods such as LEEP~\cite{nguyen2020leep} and LogME~\cite{you2021logme} implicitly favor linear separability, while embedding-based scores such as GBC~\cite{pandy2022transferability} and CCFV~\cite{yang2023pick} often assume simple parametric statistics (e.g., Gaussianity). However, segmentation quality depends less on coarse global class separation and more on whether features preserve local geometric structure near high-frequency boundaries. As shown in Fig.~\ref{fig:teaser} (right), purely statistical similarity may yield rankings inconsistent with fine-tuning outcomes, whereas a topology-aware view of the feature manifold better reflects boundary separability and thus transferability.

Our method bridges these gaps through a topology-driven approach. \textbf{Rather than forcing complex feature spaces into predefined statistical molds, we utilize non-parametric, graph-theoretic structures to quantify manifold alignment. }By explicitly modeling local boundary regions and dynamically balancing global and local priors, our framework could match the specific structural complexity of diverse downstream medical targets. Specifically, our method introduces three novel components tailored for medical segmentation, as shown in Figure~\ref{fig:framework}:
(1)~\textit{Representation Topology Divergence (GRTD)} quantifies structural alignment by measuring the discrepancy between Minimum Spanning Trees (MST) constructed in the feature and label spaces, capturing the overall manifold isomorphism.
(2)~\textit{Local Boundary-Aware Topological Consistency (LBTC)} explicitly assesses manifold separability at critical boundary patches, utilizing local MST graphs to ensure distinct decision boundaries in the presence of background heterogeneity, where segmentation typically fails.
(3)~\textit{Task-Adaptive Topological Fusion} dynamically calibrates the global and local metrics based on the target task's complexity, optimally balancing the need for broad anatomical context versus fine-grained boundary details.

We validate our framework on the large-scale OpenMind benchmark~\cite{wald2025openmind}, covering 6 diverse anatomical segmentation tasks and a model zoo consisting of 7 mainstream SSL methods pre-trained on 114,000 3D volumes. Extensive experiments demonstrate that our topology-driven approach significantly outperforms existing baselines by around \textbf{31\%} relative improvement in weighted Kendall's $\tau$, offering a robust, source-free solution for efficient model selection in the era of medical foundation models.

\section{Methodology}

\subsection{Problem Formulation and Framework Overview}
We address the problem of selecting the optimal pre-trained encoder from a SSL foundation model zoo $\mathcal{M} = \{\phi_k\}_{k=1}^{K}$ for a target segmentation task $\mathcal{D} = \{(\mathbf{x}_i, \mathbf{y}_i)\}_{i=1}^{N}$ without incurring the computational cost of fine-tuning. Let $P(\phi, \mathcal{D})$ denote the ground-truth performance (e.g., Dice score) after fine-tuning. Our objective is to derive a \textit{training-free} transferability score $T(\phi, \mathcal{D})$ that serves as a reliable proxy for $P$. 
Ideally, the score should preserve the performance ranking across the model zoo, satisfying the monotonicity condition for any pair of models $\phi_i, \phi_j \in \mathcal{M}$:
\begin{equation}
    T(\phi_i, \mathcal{D}) > T(\phi_j, \mathcal{D}) \iff P(\phi_i, \mathcal{D}) > P(\phi_j, \mathcal{D})
\end{equation}

While existing metrics typically rely on distributional statistics and neglect the complex geometry of SSL manifolds required for dense prediction, we argue that for segmentation tasks, \textit{topological tractability}, i.e., the preservation of structural connectivity or boundary separation, is a rather reliable predictor of transferability than mere statistical overlap. We propose a topology-driven framework that assesses feature-label alignment at three levels: (1) \textbf{Global Representation Topology Divergence (GRTD)}, which quantifies the structural discrepancy between feature-induced and label-induced Minimum Spanning Trees (MST); (2) \textbf{Local Boundary-Aware Topological Consistency (LBTC)}, which specifically examines the feature space distinctness at anatomical boundaries, ensuring that the pre-trained features remain separable in these critical transition zones where segmentation failures most frequently occur;  and (3) \textbf{Task-Adaptive Aggregation} strategy that dynamically weights global and local metrics based on the inherent topological complexity of the target anatomy.

\begin{figure*}[t]
    \includegraphics[width=\linewidth, height=0.45\linewidth]{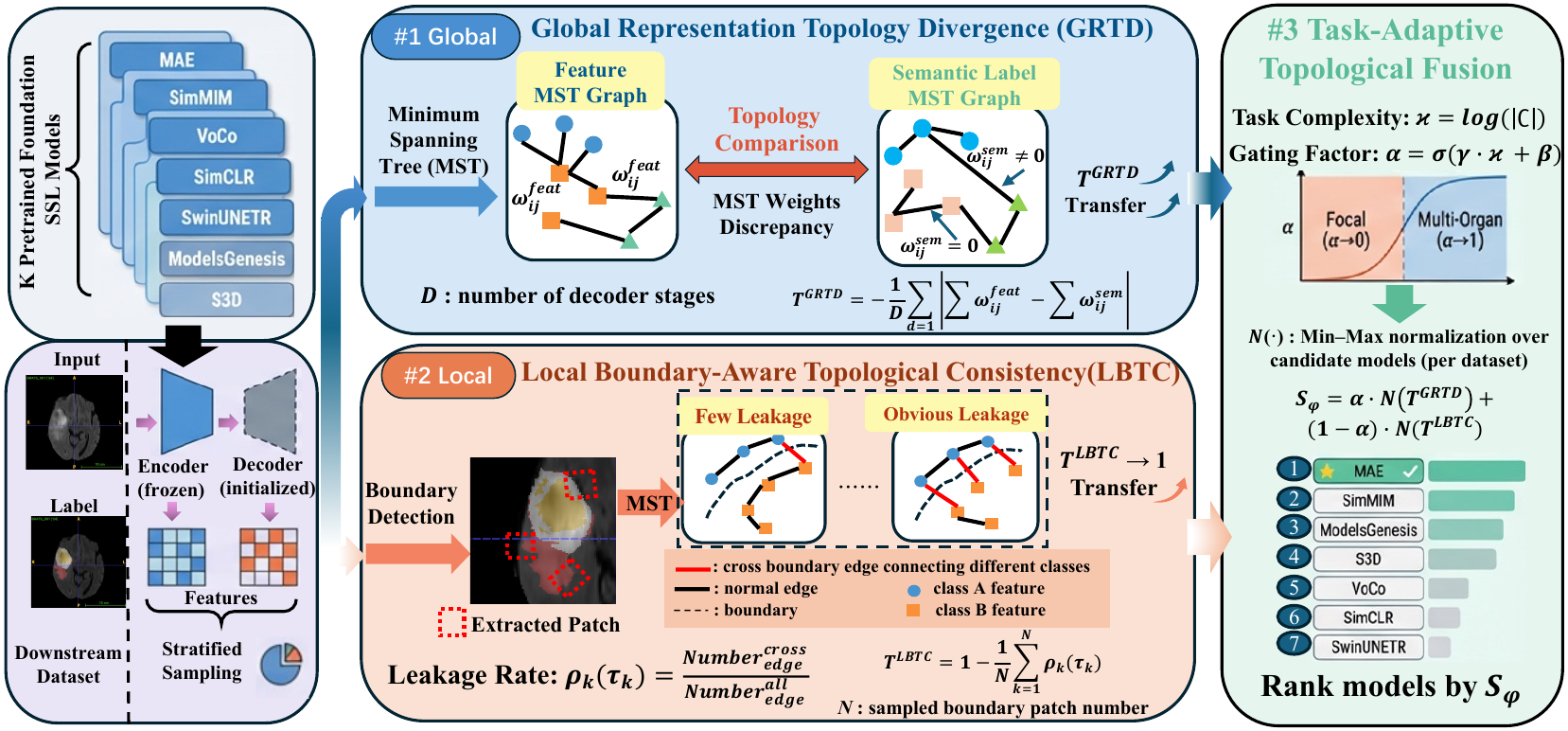}

    \caption{The overall framework. First, we extract multi-scale features via stratified sampling and construct topological graphs via MST. We then compute GRTD to quantify overall manifold alignment and LBTC to evaluate separability at critical anatomical boundaries.  These metrics are integrated via a task-complexity gating factor, producing a final score $S_{\phi}$ that predicts fine-tuning performance without training.}
    \label{fig:framework}
    
\end{figure*}

\subsection{Global Representation Topology Divergence (GRTD)}

To capture the complex, non-linear geometry without imposing parametric assumptions, we model the data distribution using the Minimum Spanning Tree (MST), which serves as a robust descriptor of the manifold's 1-skeleton, enabling us to evaluate transferability through the lens of topological isomorphism: a highly transferable model should yield a feature topology that naturally aligns with the semantic hierarchy.

Let $\mathcal{X} = \{(\mathbf{v}_i, y_i)\}_{i=1}^{N}$ be the sampled feature-label pairs. We construct two graphs to evaluate topological alignment, as shown in Figure~\ref{fig:framework}. The first is the \textit{Native Feature Graph} $\mathcal{G}_{feat}$, where edge weights represent Euclidean distances in the embedding space: $w_{ij}^{feat} = \|\mathbf{v}_i - \mathbf{v}_j\|_2$. The second is the \textit{Semantic Label-Induced Graph} \(\mathcal{G}_{sem}\), which represents an ideal label-driven topology. It explicitly injects semantic constraints by forcing samples of the same class to perfectly cluster while preserving the original feature distances for inter-class pairs up to a maximum penalty:
\begin{equation}
    w_{ij}^{sem} = 
    \begin{cases} 
    0 & \text{if } y_i = y_j, \\
    \min(\|\mathbf{v}_i - \mathbf{v}_j\|_2, \lambda) & \text{if } y_i \neq y_j.
    \end{cases}
\end{equation}
where $\lambda$ penalizes inter-class transitions. The MSTs derived from these graphs, denoted $\mathcal{T}_{feat}$ and $\mathcal{T}_{sem}$, represent the natural clustering tendency and the ideal semantic connectivity, respectively.

We define the GRTD as the discrepancy between the total weights of these two topological structures. Aggregating across $D$ decoder stages, the final score is formulated as:
\begin{equation}
    T^{\text{GRTD}} = - \frac{1}{D} \sum_{d=1}^{D} \left| \sum_{(i,j) \in \mathcal{T}_{feat}^d} w_{ij}^{feat} - \sum_{(i,j) \in \mathcal{T}_{sem}^d} w_{ij}^{sem} \right|.
    \label{eq:rtd}
\end{equation}
A higher $T^{\text{GRTD}}$ (closer to 0) indicates that the encoder's native geometry naturally respects semantic boundaries.

\subsection{Local Boundary-Aware Topological Consistency (LBTC)}

Although GRTD captures the global manifold topology, it is susceptible to the class imbalance typical of medical images, where low-frequency background regions dominate the feature space statistics. However, the success of segmentation often hinges on the model's ability to preserve high-frequency details at critical anatomical boundaries. Hence, we propose the Local Boundary-Aware Topological Consistency (LBTC), shifting the focus from global isomorphism to local separability, explicitly scrutinizing whether the pre-trained features maintain distinct decision boundaries within ambiguous boundary patches.

We first define the set of boundary anchors $\partial \mathcal{Y}$ via the morphological gradient of the ground truth masks. For each anchor $c_k \in \partial \mathcal{Y}$, we extract a local patch $\mathcal{P}_k$ and construct a local graph $\mathcal{G}_k$. The Minimum Spanning Tree of this local neighborhood, $\mathcal{T}_k$, serves as a probe for local cluster purity. Ideally, $\mathcal{T}_k$ should traverse all intra-class nodes before bridging the semantic gap. We quantify the \textit{Topological Leakage Rate} $\rho_k$, which measures the proportion of edges in the local MST that erroneously connect distinct semantic classes:
\begin{equation}
    \rho_k(\mathcal{T}_k) = \frac{1}{|\mathcal{V}_k|-1} \sum_{(u,v) \in \mathcal{E}(\mathcal{T}_k)} \mathbb{I}(y_u \neq y_v),
\end{equation}
where $\mathcal{E}(\mathcal{T}_k)$ denotes the edge set of the local MST and $\mathbb{I}(\cdot)$ is the indicator function. The final LBTC score is derived by aggregating these local inconsistencies over $N$ sampled boundary regions, formulated as the complement of the expected leakage:
\begin{equation}
    T^{\text{LBTC}} = 1 - \frac{1}{N} \sum_{k=1}^{N} \rho_k(\mathcal{T}_k).
    \label{eq:lbtc}
\end{equation}
A score approaching 1 implies that the encoder preserves strict topological separation even within the ambiguous transition zones of the manifold.

\subsection{Task-Adaptive Topological Fusion}

Medical segmentation targets exhibit distinct topological regimes: multi-organ tasks demand global structural preservation (high structural complexity), while small lesion extraction relies on local boundary contrast (high boundary complexity). A static combination of global and local metrics fails to generalize across these heterogeneous distributions. To resolve this, we propose a dynamic fusion mechanism driven by the semantic cardinality of the target task. 

We define a task complexity prior $\kappa = \log(|\mathcal{C}|)$, where $|\mathcal{C}|$ is the number of semantic classes. This prior modulates a gating factor $\alpha \in (0, 1)$ via a sigmoid function $\sigma(\cdot)$, shifting the focus between macro-structure and micro-details. Let $\mathcal{N}(\cdot)$ denote the Min-Max normalization operator over the model zoo. The final transferability score $\mathcal{S}_{\phi}$ is computed as a convex combination of the global and local topological metrics:
\begin{equation}
    \alpha = \sigma(\gamma \cdot \kappa + \beta), \quad 
    \mathcal{S}_{\phi} = \alpha \cdot \mathcal{N}(T^{\text{GRTD}}) + (1 - \alpha) \cdot \mathcal{N}(T^{\text{LBTC}}),
    \label{eq:final_score}
\end{equation}
where $\gamma$ and $\beta$ are scaling constants. This formulation ensures that for complex anatomical tasks ($\alpha \to 1$), the ranking prioritizes global layout isomorphism, whereas for focal pathologies ($\alpha \to 0$), it emphasizes the sharpness of local decision boundaries.

\begin{table*}[tp]
\centering
\caption{Weighted Kendall’s $\tau$ for transferability estimation on the OpenMind benchmark.}
\label{tab:main_results}
\footnotesize
\renewcommand{\arraystretch}{1.05}
\setlength{\tabcolsep}{4.2pt}
\begin{tabular}{l c c c c | c c | c}
\toprule
\multirow{2}{*}{\textbf{Method}} &
\multicolumn{4}{c}{\textbf{ID (Same Region)}} &
\multicolumn{2}{c}{\textbf{OOD (Different Region)}} &
\textbf{Avg} \\
\cmidrule(lr){2-5} \cmidrule(lr){6-7}
& \textbf{MSF} & \textbf{ISL} & \textbf{HNT} & \textbf{TPC} & \textbf{ACD} & \textbf{KIT} & \textbf{All} \\
\midrule
LogME~\cite{you2021logme}
 & -0.524 & -0.223 & -0.315 & -0.545 & 0.575 & -0.649 & -0.280 \\
\midrule
LEEP~\cite{nguyen2020leep}
 & -0.220 & -0.427 & -0.492 & -0.051 & -0.621 & 0.225 & -0.264 \\
\midrule
GBC~\cite{pandy2022transferability}
 & -0.707 & -0.601 & -0.537 & -0.256 & -0.325 & -0.621 & -0.508 \\
\midrule
CCFV~\cite{yang2023pick}
 & 0.277 & 0.503 & 0.869 & 0.664 & \textbf{0.817} & 0.180 & 0.552 \\
\midrule
Ours
 & \textbf{0.705} & \textbf{0.814} & \textbf{0.905} & \textbf{0.664} & 0.671 & \textbf{0.578} & \textbf{0.723} \\
\bottomrule
\end{tabular}
\end{table*}

\section{Experiment and Results}

\subsection{The Foundation Model Zoo and Downstream Tasks}
We utilize the OpenMind benchmark~\cite{wald2025openmind} model zoo, featuring ResEnc-L~\cite{isensee2024nnu} backbones pre-trained on 114,000 unlabeled 3D brain MRIs. To rigorously evaluate transferability without prior label exposure, we select models trained via reconstruction (MAE\cite{He2022MAE}, SimMIM\cite{chen2023masked}, ModelsGenesis\cite{zhou2021models}, S3D~\cite{wald2025revisiting}) and contrastive learning (VoCo\cite{Wu2024Voco}, SimCLR\cite{chen2020simple}, SwinUNETR\cite{Tang2022SwinUNETR}). We evaluate these models across diverse downstream tasks spanning various anatomies and modalities.

\noindent
\textbf{In-Distribution (ID) Tasks:} We select four datasets closely aligned with the pre-training domain (head-and-neck) to test intrinsic transferability: \textit{ISLES (ISL)}~\cite{hernandez2022isles} (stroke lesions on DWI/ADC), \textit{HNTS-MRG (HNT)}~\cite{wahid2024hntsmrg} (head-and-neck tumors on MR), \textit{MS FLAIR (MSF)}~\cite{muslim2022ms} (multiple-sclerosis lesions), and \textit{ToP-CoW (TPC)}~\cite{yang2024topcow} (Circle of Willis vessels).
\textbf{Out-of-Distribution (OOD) Tasks:} To assess generalization under distribution shifts, we include two OOD tasks: \textit{ACDC (ACD)}~\cite{bernard2018acdc} (cardiac structures on cine-MRI), and \textit{KiTS19 (KIT)}~\cite{heller2021state} (kidney and tumors on CT). Notably, KIT represents a challenging cross-anatomy and cross-modality (MR$\rightarrow$CT) transfer scenario with clear tumor boundaries~\cite{isensee2024nnu}.

\begin{figure*}[tbp]
    \includegraphics[width=0.99\linewidth, height=0.3\linewidth]{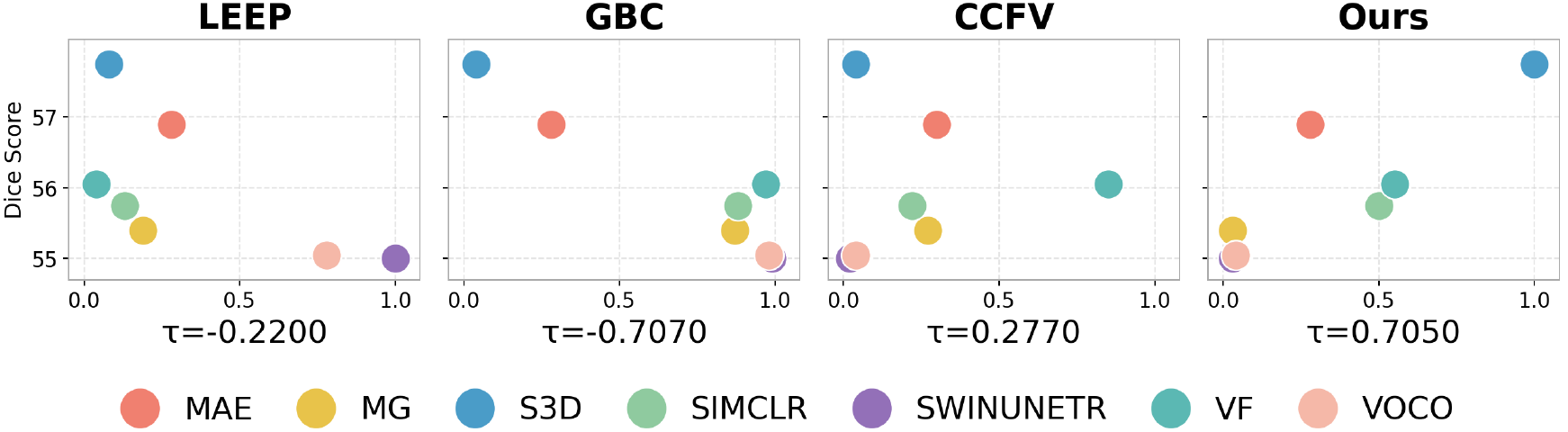} 

    \caption{Correlation between the fine-tuning performance and transferability metrics using MSF as an example. The vertical axis represents the average Dice, while the horizontal axis represents the \emph{standardized} transferability metric. We want to observe a positive relationship between higher performance and higher transferability estimations.}
\label{fig:correlation_matrix}
    
\end{figure*}

\subsection{Implementation Details}
\label{subsec:impl}

Considering the SSL foundation model just have well-trained encoders, we attach a randomly initialized nnU-Net~\cite{isensee2021nnu} decoder to each candidate model, extracting multi-scale features via sliding-window inference. To ensure topological fidelity despite severe class imbalance, we employ a stratified sampling strategy that balances class-conditional foreground voxels against background and global distributions. We differentiate the extraction depth to align with topological roles: GRTD is computed on the final $k$ decoder layers to capture semantic layout, while Local LBTC is evaluated on the last $k$ encoder layers to assess intrinsic boundary separability. The task-adaptive gating weights are estimated using a computationally efficient pilot set of $K{=}10$ cases, with final rankings quantified via the weighted Kendall correlation ($\tau^{*,w}$)~\cite{vigna2015weighted}.

\subsection{Results and Analysis}

\noindent
\textbf{Overall Performance}
We report the weighted Kendall’s \(\tau\)  between the estimated transferability scores and the actual fine-tuning performance. As shown in Table \ref{tab:main_results} and Figure~\ref{fig:correlation_matrix}, traditional classification-based metrics (LogME, LEEP) exhibit severe negative correlations, confirming their inadequacy for dense prediction tasks. While the embedding-based CCFV shows positive correlations, it struggles on challenging out-of-distribution (OOD) transfers like KIT (\(\tau=0.180\)). In contrast, our method demonstrates robust and consistent performance across both in-distribution and OOD scenarios, achieving the highest average correlation of 0.723, validating that topological tractability is a superior proxy for segmentation transferability.

\noindent
\textbf{Robustness to Decoder Initialization.}
Although the decoder is randomly initialized, our metric should primarily reflect the intrinsic quality of the \emph{pre-trained} representation rather than stochasticity from initialization.
As summarized in Table~\ref{tab:init_time} (left), the weighted Kendall's $\tau^{*}_{w}$ remains consistently stable across Kaiming~\cite{he2015delving}/Xavier~\cite{glorot2010understanding}/Gaussian initializations with small variance, suggesting that the topology-driven signal dominates over initialization noise.

\noindent
\textbf{Computational Time Comparison.}
\label{sec:time}
Table~\ref{tab:init_time} (right) contrasts the practical cost of model selection via training-free scoring versus exhaustive downstream fine-tuning.
Compared with CCFV, our approach substantially reduces the metric-computation overhead, enabling rapid screening over a model zoo.
More importantly, it avoids the prohibitive end-to-end cost of repeating fine-tuning across multiple candidates, making transferability estimation a realistic substitute for brute-force selection in resource-constrained settings.

\noindent
\textbf{Impact of Target Topology}
Table~\ref{tab:ablation} reveals the complementary nature of our topological priors. While Global (GRTD) and Local (LBTC) metrics suffer significant performance drops on fragmented and structured targets respectively, our adaptive fusion effectively bridges this gap and achieves a robust average of \textbf{0.714}, a substantial improvement over the single-stream baselines.

\begin{table*}[t]
\centering
\caption{ \textbf{Left:} weighted Kendall's $\tau^{*}_{w}$ averaged over five random seeds for each decoder initialization scheme.
\textbf{Right:} transferability time and fine-tuning time are averaged over five repeated timing trials; both are reported as the \emph{total} wall-clock time for all 7 models. \emph{All times are in minutes.}}
\label{tab:init_time}

\footnotesize
\renewcommand{\arraystretch}{1.10}
\setlength{\tabcolsep}{4.2pt}

\begin{minipage}[t]{0.49\textwidth}
\centering
\textbf{Decoder initialization robustness}\\
\begin{tabular}{lccc}
\toprule
Dataset & Kaiming & Xavier & Gaussian \\
\midrule
TPC & 0.664 & 0.646 & 0.652 \\
ACD & 0.671 & 0.624 & 0.652 \\
MSF & 0.705 & 0.706 & 0.718 \\
\midrule
Avg. & 0.680 & 0.659 & 0.674 \\
\bottomrule
\end{tabular}
\end{minipage}
\hfill
\begin{minipage}[t]{0.49\textwidth}
\centering
\textbf{Time comparison}\\
\begin{tabular}{lccc}
\toprule
Dataset & CCFV & Ours & Fine-tune \\
\midrule
TPC & 1050.5 & \textbf{10.07} & 3000$+$ \\
ACD & 756.9 & \textbf{8.02} & 3000$+$ \\
MSF & 22.3 & \textbf{2.89} & 3000$+$ \\
\midrule
Avg. & 609.9 & \textbf{6.99} & 3000$+$ \\
\bottomrule
\end{tabular}
\end{minipage}
\end{table*}

\begin{table}[tp]
\centering
\caption{Ablation study on topological priors of different datasets. We stratify tasks into \textit{Fragmented} (ISL, MSF) and \textit{Structured} (ACD, TPC) targets. Our \textbf{Adaptive} fusion achieves robust performance across all regimes.}
\label{tab:ablation}
\setlength{\tabcolsep}{4pt}
\renewcommand{\arraystretch}{1.05}
\begin{tabular}{l|cc|cc|c}
\toprule
\textbf{Method} & \textbf{ISL} & \textbf{MSF} & \textbf{ACD} & \textbf{TPC} & \textbf{Avg.} \\
\midrule
Global (GRTD) & 0.122 & 0.333 & 0.624 & \textbf{0.664} & 0.436 \\
Local (LBTC)  & \textbf{0.814} & \textbf{0.705} & 0.319 & 0.062 & 0.475 \\
\midrule
\textbf{Adaptive (Ours)} & \textbf{0.814} & \textbf{0.705} & \textbf{0.671} & \textbf{0.664} & \textbf{0.714} \\
\bottomrule
\end{tabular}
\end{table}

\section{Conclusion}
In this paper, we present a novel topology-driven transferability estimation framework of medical foundation model for segmentation. By shifting the evaluation paradigm from statistical overlap to manifold topology, our approach uniquely captures both global structural isomorphism and local boundary separability. Extensive validation on the OpenMind benchmark demonstrates that our task-adaptive method significantly outperforms existing baselines across diverse in-distribution and out-of-distribution tasks. By providing a robust, training-free proxy for model selection, our framework eliminates the prohibitive computational costs of exhaustive fine-tuning, paving the way for the efficient and scalable clinical deployment of medical foundation models.

%
%
\bibliographystyle{splncs04}
\bibliography{cite}

\end{document}